%% file: main.tex
\documentclass[a4paper]{article}

\usepackage{INTERSPEECH2022}
\usepackage{hyperref}
\hypersetup{colorlinks=true}
\usepackage{enumitem}
\usepackage{multirow}
\usepackage{bm}
\usepackage{IEEEtrantools}
\usepackage[bb=dsserif]{mathalpha}

\DeclareGraphicsExtensions{.pdf,.png}

\title{Online Model Compression for Federated Learning with Large Models}
\name{Tien-Ju Yang, Yonghui Xiao, Giovanni Motta, Françoise Beaufays, Rajiv Mathews, Mingqing Chen}
\address{Google LLC, Mountain View, CA, U.S.A.}
\email{\{tjy,yohu,giovannimotta,fsb,mathews,mingqing\}@google.com}

\begin{document}

\bstctlcite{IEEEexample:BSTcontrol}

\maketitle


\input{0_abstract}


\input{1_introduction}

\input{2_methodology}

\input{3_experimental_results}

\input{4_related_works}

\input{5_conclusion}


\clearpage
{
\bibliographystyle{IEEEtran}
\input{__references.bbl}

}

\input{6_acknowledgements}

\end{document}

%% file: 0_abstract.tex
\begin{abstract}

This paper addresses the challenges of training large neural network models under federated learning settings: high on-device memory usage and communication cost. The proposed Online Model Compression (OMC) provides a framework that stores model parameters in a compressed format and decompresses them only when needed. We use quantization as the compression method in this paper and propose three methods, (1) using per-variable transformation, (2) weight matrices only quantization, and (3) partial parameter quantization, to minimize the impact on model accuracy. According to our experiments on two recent neural networks for speech recognition and two different datasets, OMC can reduce memory usage and communication cost of model parameters by up to 59\% while attaining comparable accuracy and training speed when compared with full-precision training.

\end{abstract}

%% file: 1_introduction.tex
\section{Introduction}

Federated learning (FL)~\cite{kairouz_2019_openproblem,wang_2021_fieldguide} allows training neural network models directly on edge devices (referred to as \emph{clients}) instead of transferring their data back to a server for centralized training to preserve users' privacy. FL is composed of multiple \emph{federated rounds}. In a standard federated round, a server model is first transported to clients. Then, the clients train the model on their local data. The trained models are finally transported back to the server and aggregated to improve the server model. This process is repeated until the server model converges.

FL involves on-device training and model transportation between servers and clients, which lead to two main challenges. The first challenge is that edge devices usually have limited memory available for training. Given the fact that recent Automatic Speech Recognition (ASR) models typically contain hundreds of millions of parameters or more~\cite{gulati_2020_conformer}, keeping these parameters in full precision in memory may exceed the available memory. Although there is a significant effort in the field on reducing memory usage of parameters during \emph{inference}, such as quantization-aware training~\cite{rastegari_2016_xnornet,abdolrashidi_2021_quantized_resnet}, it is usually at the cost of higher memory usage during training. Reducing the memory usage of parameters during \emph{training} with FL is less explored. The second challenge is the high communication cost. Communication can be much slower than computation~\cite{huang_2013_communication_speed}, and transporting models in full precision also burdens the communication network.

In this paper, we propose \emph{Online Model Compression (OMC)} to address the above challenges of on-device FL. Different from regular full-precision FL, where each client keeps, updates, and transports \emph{full-precision} parameters, OMC keeps and transports the parameters in a compressed format. During training, when an operation needs the value of a compressed parameter, OMC decompresses it on-the-fly and deallocates memory for the decompressed value immediately after it is consumed. Therefore, OMC only keeps the compressed parameters and a small number of transient decompressed parameters in memory, which uses less memory than the full-precision parameters.

The main design challenge of OMC is achieving a favorable accuracy-efficiency trade-off. An important characteristic of OMC is that compression and decompression occur in every training iteration. As a result, the error introduced by compression can accumulate very quickly and degrade model accuracy significantly. On the other hand, we cannot use a very complicated algorithm to control the accumulated error because this will significantly slow down training. Therefore, OMC needs to be as simple and fast as possible and has a minimal impact on accuracy. It achieves this goal by using quantization, per-variable transformation, weight matrices only quantization, and partial parameter quantization.

The following summarizes the benefits of OMC:
\begin{itemize}[nolistsep]
    \item \textbf{Reducing memory usage}: There are three main sources of memory usage: model parameters, activations, and gradients. OMC aims to reduce the memory usage of model parameters.
    \item \textbf{Reducing communication cost}: Because models are transported between servers and clients, reducing the model size helps reduce communication cost.
    \item \textbf{Lightweight operation}: OMC does not significantly slow down the training process even though compression and decompression occur frequently.
\end{itemize}

%% file: 2_methodology.tex
\section{Methodology}
\label{sec:methodology}

\subsection{Framework of Online Model Compression}

\begin{figure}[t]
    \centering
    \includegraphics[width=0.47\textwidth]{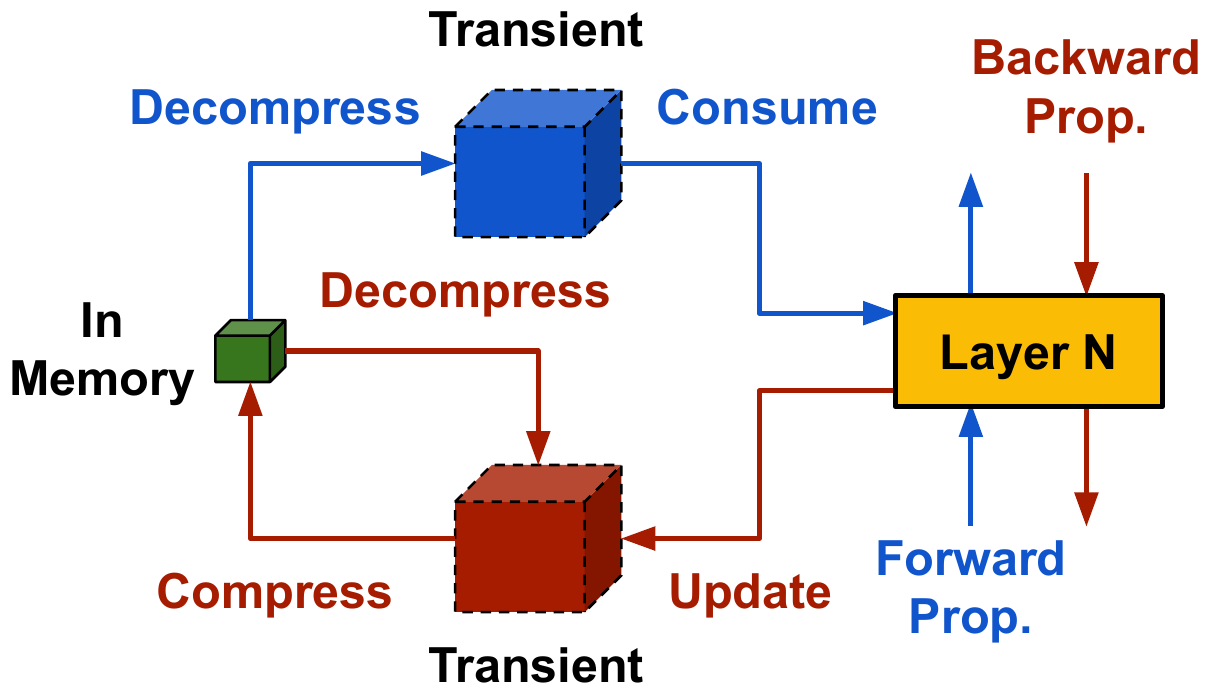}
    \caption{The illustration of the framework of the proposed online model compression. The cubes with dashed borderlines are transient variables.}
    \vspace{-5pt}
    \label{fig:omc_framework}
\end{figure}

Fig.~\ref{fig:omc_framework} illustrates the framework of the proposed Online Model Compression (OMC). OMC stores parameters in a compressed format, such as floating-point numbers with reduced bitwidths, but performs computations in full precision or other hardware-supported formats. This design decouples compression formats and hardware-supported formats to provide higher flexibility for choosing the compression format and method to achieve better memory usage reduction.

When performing forward propagation for a layer (the blue path in Fig.~\ref{fig:omc_framework}), OMC decompresses the required parameters for that layer on the fly and deallocates the decompressed copies immediately after they are consumed. When performing backward propagation for a layer (the red path in Fig.~\ref{fig:omc_framework}), OMC decompresses the required parameters and applies the gradients to update them. The updated decompressed parameters are then compressed and discarded immediately. Therefore, OMC only keeps the compressed parameters and a small number of transient decompressed copies in memory.

\subsection{Quantization-Based Online Model Compression}

Given the simplicity of quantization, we adopt it as the compression method in this paper. Quantization reduces the number of bits (i.e., bitwidth) for representing a value. While full precision (32 bits) is usually used in deep learning, many works in the literature have shown that neural networks are error-resilient and allow using much lower bitwidths, such as 1 to 8 bits, without harming prediction accuracy~\cite{abdolrashidi_2021_quantized_resnet}. However, such low bitwidths are usually achieved for inference. Reducing memory usage by quantization during training is more difficult because training requires more bits to precisely accumulate the small gradients across training iterations.

OMC adopts the floating-point format in this paper as an example although other formats, such as the fixed-point format, can also be used. The floating-point format consists of three parts: the sign bits, the exponent bits, and the mantissa bits. For example, the format of FP32 (32-bit single-precision floating-point format) is composed of 1-bit sign, 8-bit exponent, and 23-bit mantissa. To quantize a floating-point value, we reduce the numbers of bits for the exponent and the mantissa, which are the two hyper-parameters of floating-point quantization.

\subsection{Per-Variable Transformation}

\begin{figure}[t]
    \centering
    \includegraphics[width=0.47\textwidth]{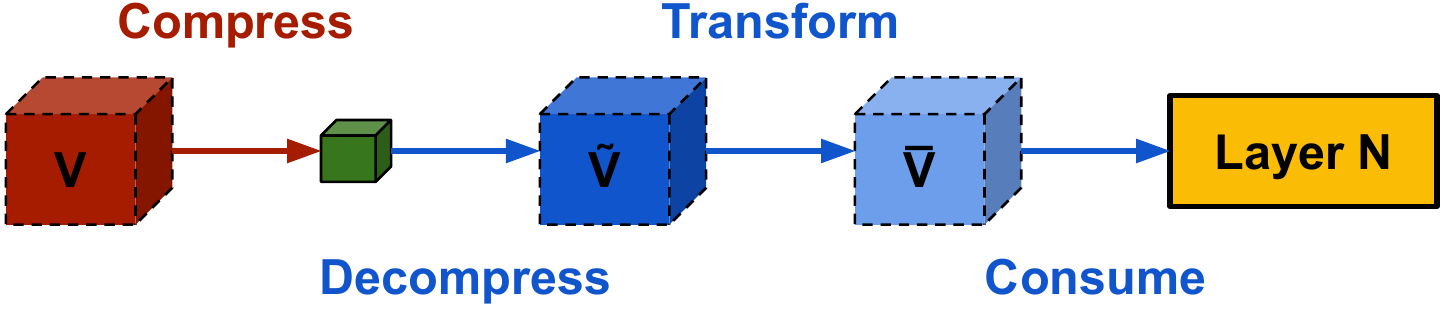}
    \caption{The illustration of per-variable transformation. The cubes with dashed borderlines are transient variables.}
    \vspace{-5pt}
    \label{fig:per_variable_transformation_illustration}
\end{figure}

Quantization is a lossy operation and thus, introduces quantization errors. As a result, quantizing parameters every training iteration can lead to a large accumulated error and prevent us from using fewer bits with the original accuracy maintained. To minimize the quantization error, OMC applies a linear transformation on the decompressed parameters, which is illustrated in Fig.~\ref{fig:per_variable_transformation_illustration}. This step is performed \emph{per variable}, such as per weight matrices, so that all the model parameters in a variable can share a few transformation-related parameters to make the memory overhead negligible.

The transformed variable (vector or flattened tensor) $\boldsymbol{\bar{V}}\in\mathbb{R}^{n}$ can be written as $\boldsymbol{\bar{V}}=s\boldsymbol{\tilde{V}} + b\boldsymbol{\mathbb{1}}$, where $\boldsymbol{\tilde{V}}\in\mathbb{R}^{n}$ denotes the decompressed variable, $\boldsymbol{\mathbb{1}}\in\mathbb{R}^{n}$ is a one vector, and $s$ and $b$ denote the \emph{per-variable} scaling factor and bias, respectively. OMC determines the scaling factor and the bias analytically by minimizing the $\ell2$-norm of the difference between the decompressed and transformed variable ($\boldsymbol{\bar{V}}$) and the full-precision variable before compression ($\boldsymbol{V}\in\mathbb{R}^{n}$). The closed-form solutions are
\begin{align*}
s=&\frac{n\sum_{k}V_k\tilde{V}_k-\sum_{k}V_k\sum_{k}\tilde{V}_k}{n\sum_{k}V_k^2-(\sum_{k}\tilde{V}_k)^2}, \\
b=&\frac{n\sum_{k}V_k-s\sum_{k}\tilde{V}_k}{n},
\end{align*}
where $V_k$ denotes the $k$-th element of $\boldsymbol{V}$. Note that the degenerated case is when the denominator of $s$ is $0$, which happens (only) when all elements in each of $\boldsymbol{V}$ and $\boldsymbol{\tilde{V}}$ are the same according to inequality of arithmetic and geometric means. We set $s$ to $1.0$ for this degenerated case. In our implementation, $s$ and $b$ are computed in the 64-bit floating-point precision, but the final $s$ and $b$ are still stored as FP32 values.

\subsection{Weight Matrices Only Quantization}

We empirically found that some types of parameters are more sensitive to quantization than the others. These sensitive parameters include the scaling factors and biases in normalization layers. In contrast, weight matrices in convolutional and feed-forward layers are less sensitive to quantization but dominate the model size. For example, the weight matrices in the streaming Conformer model we use in Sec.~\ref{sec:experimental_results} accounts for 99.8\% of the model size. Hence, OMC only quantizes weight matrices and keeps the remaining variables in FP32. This method helps maintain accuracy while saving a large amount of memory.

\subsection{Partial Parameter Quantization}

OMC also leverages the feature of federated learning that there are many clients training a model in parallel to further reduce quantization errors. This feature provides an opportunity to quantize only a subset of parameters for each client and vary the selection from one client to another. As a result, the server can receive high-quality and precise update of each parameter from the clients that do not quantize this parameter.

%% file: 3_experimental_results.tex
\section{Experimental Results}
\label{sec:experimental_results}

\begin{table*}[t]
\centering
\scalebox{1.0}{
\begin{tabular}{c|c|ccc}
\toprule
\multirow{2}{*}{}          & \multirow{2}{*}{WERs}          & \multicolumn{2}{c}{Resource}                        \\ \cline{3-4} 
                           &                                & Parameter Memory / Communication   & Speed (Rounds/Min)   \\ \hline
FP32 (S1E8M23)             & 2.1/4.6/2.2/4.8                & 474MB (100\%)                & 29.5 (100\%)         \\
\textbf{OMC (S1E4M14)}     & \textbf{2.1/4.7/2.2/4.6}       & \textbf{301MB (64\%)}        & \textbf{26.8 (91\%)} \\
\bottomrule
\end{tabular}
}
\caption{The results of Non-Streaming Conformer on IID LibriSpeech.}
\vspace{-12pt}
\label{table:ns_cf_iid_libri}
\end{table*}

\begin{table*}[t]
\centering
\scalebox{1.0}{
\begin{tabular}{c|c|ccc}
\toprule
\multirow{2}{*}{}          & \multirow{2}{*}{WERs}          & \multicolumn{2}{c}{Resource}                        \\ \cline{3-4} 
                           &                                & Parameter Memory / Communication   & Speed (Rounds/Min)   \\ \hline
Before Adaptation          & 6.7                            & -                            & -                    \\
FP32 (S1E8M23)             & 4.6                            & 548MB (100\%)                & 11.9 (100\%)         \\ \hline
\textbf{OMC (S1E3M7)}      & \textbf{4.6}                   & \textbf{224MB (41\%)}        & \textbf{11.1 (93\%)} \\
\textbf{OMC (S1E2M3)}      & \textbf{5.9}                   & \textbf{147MB (29\%)}        & -                    \\
\bottomrule
\end{tabular}
}
\caption{The results of Streaming Conformer on the Multi-Domain dataset. The WER is on the MF domain.}
\vspace{-10pt}
\label{table:s_cf_vs}
\end{table*}

\subsection{Experimental Settings}

In this section, we validate and demonstrate the effectiveness of OMC across various use cases, including small and large datasets, IID and non-IID data distributions, streaming and non-streaming network architectures, and from-scratch and domain-adaptation training settings.

The first dataset is the LibriSpeech dataset~\cite{panayotov_2015_librispeech}. By partitioning LibriSpeech in two different ways, we derive the \emph{IID LibriSpeech} and the \emph{Non-IID LibriSpeech} dataset from the original LibriSpeech dataset to simulate different client data distributions. \emph{IID LibriSpeech} is generated by random partition while \emph{Non-IID LibriSpeech} is generated by partitioning by speakers. For LibriSpeech related experiments, the models are trained from scratch. The Word Error Rates (WERs) will be reported in the format of \emph{dev/dev-other/test/test-other}, where each item corresponds to the WER of the dev, dev-other, test, and test-other set from left to right.

The second dataset is an anonymized Multi-Domain (MD) dataset and much larger than LibriSpeech. The MD dataset contains around 400K hours of utterances from domains such as YouTube, farfield, search, and telephony~\cite{narayanan_2019_mddataset,misra_2021_mddataset}. We partition this dataset into the Medium Form (MF) domain dataset and the Non-MF domain dataset. These two partitions will be used to evaluate OMC under the domain adaptation scenario (from Non-MF domain to MF domain). For MD related experiments, a model will be first trained on the Non-MF domain dataset and then finetuned on the MF domain dataset. The WERs are reported on a disjoint test set from the MF domain.

We also experiment with two ASR models to evaluate OMC under non-streaming and streaming use cases. The first model is similar to the largest Conformer in the paper~\cite{gulati_2020_conformer}. The only difference is that we replace batch normalization by group normalization, which is more suitable for federated learning at a small degradation in accuracy~\cite{hsieh_2019_decentralized_ml}. We refer to this model as \emph{Non-streaming Conformer}. The second model is our production-grade Conformer variant~\cite{li_2021_streaming_conformer}, which contains approximately 130M trainable parameters and supports streaming use cases. We refer to this model as \emph{streaming Conformer}.

Unless otherwise specified, we randomly quantize 90\% of the weight matrices and vary the selection from round to round and from client to client. There are 128 clients, and each client trains a model with 1 local step. The batch size is 16 per client. For resource consumption, we report the theoretical memory usage of parameters, the communication cost, and the training speed on TPUs. The memory saving observed in practice with our implementation is also provided in Sec.~\ref{subsubsec:memory_consumption_on_pixel_phones}. We use \emph{SxEyMz} to represent a floating-point format with $x$ sign bits, $y$ exponent bits and $z$ mantissa bits. For example, the FP32 format is represented by S1E8M23.

\subsection{Non-Streaming Conformer on LibriSpeech}

Table~\ref{table:ns_cf_iid_libri} summarizes the results of Non-Streaming Conformer on IID LibriSpeech. Compared with FP32 (S1E8M32), OMC can achieve similar WERs with 64\% memory usage of parameters and communication cost by using the 19-bit S1E4M14 format. OMC is also pretty lightweight. In this experiment, OMC only decreases the speed by 9\%.

Table~\ref{table:ns_cf_noniid_libri} summarizes the WERs of Non-Streaming Conformer on Non-IID LibriSpeech with the same bitwidth as that of the IID LibriSpeech experiment. Even with non-IID data, OMC can still attain comparable WERs to using FP32. The reduction in memory usage of parameters and communication cost is the same as the previous IID experiment and, hence, omitted in Table~\ref{table:ns_cf_noniid_libri}. These experiments show the versatility of OMC to work well with both IID and non-IID data distribution.

\begin{table}[t]
\centering
\scalebox{1.0}{
\begin{tabular}{c|cc}
\toprule
    & FP32 (S1E8M23)             & \textbf{OMC (S1E4M14)}             \\ \hline
WER & 2.0/4.7/2.2/4.9 & \textbf{2.0/4.8/2.2/4.9} \\
\bottomrule
\end{tabular}
}
\caption{The WERs of Non-Streaming Conformer on Non-IID LibriSpeech.}
\vspace{-10pt}
\label{table:ns_cf_noniid_libri}
\end{table}

\subsection{Streaming Conformer on Multi-Domain Dataset}

We observe that domain adaptation may allow using a smaller bitwidth than from-scratch training. Table~\ref{table:s_cf_vs} summarizes the results of Streaming Conformer on the Multi-Domain dataset. Compared to FP32, OMC can achieve similar WERs with 41\% memory usage of parameters and communication cost by using the 11-bit S1E3M7 format. We can further reduce the bitwidth to 6 bits (S1E2M3) and still improve the WERs over the before-adaptation baseline. Moreover, OMC only has a negligible impact on the training speed, by 7\% in this case.

\subsection{Measured Memory Usage on Pixel 4 Phones}
\label{subsubsec:memory_consumption_on_pixel_phones}

In this section, we measured the memory usage on Google Pixel 4 with parameters quantized to FP16 (S1E5M10). We implemented federated learning with Tensorflow Federated~\cite{tff_url} and applied gradient recomputation~\cite{chen_2016_gradient_recomputation} to force releasing the memory occupied by transient parameters. The code has been uploaded to the Lingvo~\cite{DBLP:journals/corr/abs-1902-08295, lingvo_url} repository on Github. For the Streaming Conformer, OMC reduces the peak memory usage by 197 MB (38\% of the model size). For a smaller Streaming Conformer model with 3 Conformer blocks in the encoder, OMC reduces the peak memory usage by 84 MB (45\% of the model size).

\subsection{Ablation Study}

In the ablation study, we use Streaming Conformer on Multi-Domain dataset as the study target unless otherwise specified.

\subsubsection{Impact of Proposed Methods}

We start from studying the impact of each of the proposed methods on WERs. The results are summarized in Table~\ref{table:wer_change}. After we quantize the parameters to 11 bits (S1E3M7), the WER significantly increases by 2.3. The WER gap is first closed by the proposed per-variable transformation, which reduces the WER by 0.4. Then, the proposed weight matrices only quantization reduces the WER by another 1.8. Finally, the proposed partial parameter quantization brings down the WER to 4.6, which matches the FP32 baseline.

\begin{table}[t]
\centering
\scalebox{0.85}{
\begin{tabular}{c|c|c|c||c}
\toprule
\begin{tabular}[c]{@{}c@{}}Quantization\\(S1E3M7)\end{tabular}   & \begin{tabular}[c]{@{}c@{}}Per-Variable\\Transformation\end{tabular} & \begin{tabular}[c]{@{}c@{}}Weights\\Only\end{tabular} & \begin{tabular}[c]{@{}c@{}}90\%\\Weights\end{tabular} & WER \\ \toprule
           &                                                         &                                                        &                                                        & 4.6 \\ \hline
\checkmark &                                                         &                                                        &                                                        & 6.9 \\ \hline
\checkmark & \checkmark                                              &                                                        &                                                        & 6.5 \\ \hline
\checkmark & \checkmark                                              & \checkmark                                             &                                                        & 4.7 \\ \hline
\checkmark & \checkmark                                              & \checkmark                                             & \checkmark                                             & 4.6 \\ \toprule
\end{tabular}
}
\caption{The change in WER when we apply each of the proposed methods sequentially.}
\vspace{-10pt}
\label{table:wer_change}
\end{table}

\subsubsection{Per-Variable Transformation for From-Scratch Training}

\begin{figure}[t]
    \centering
    \includegraphics[width=0.35\textwidth]{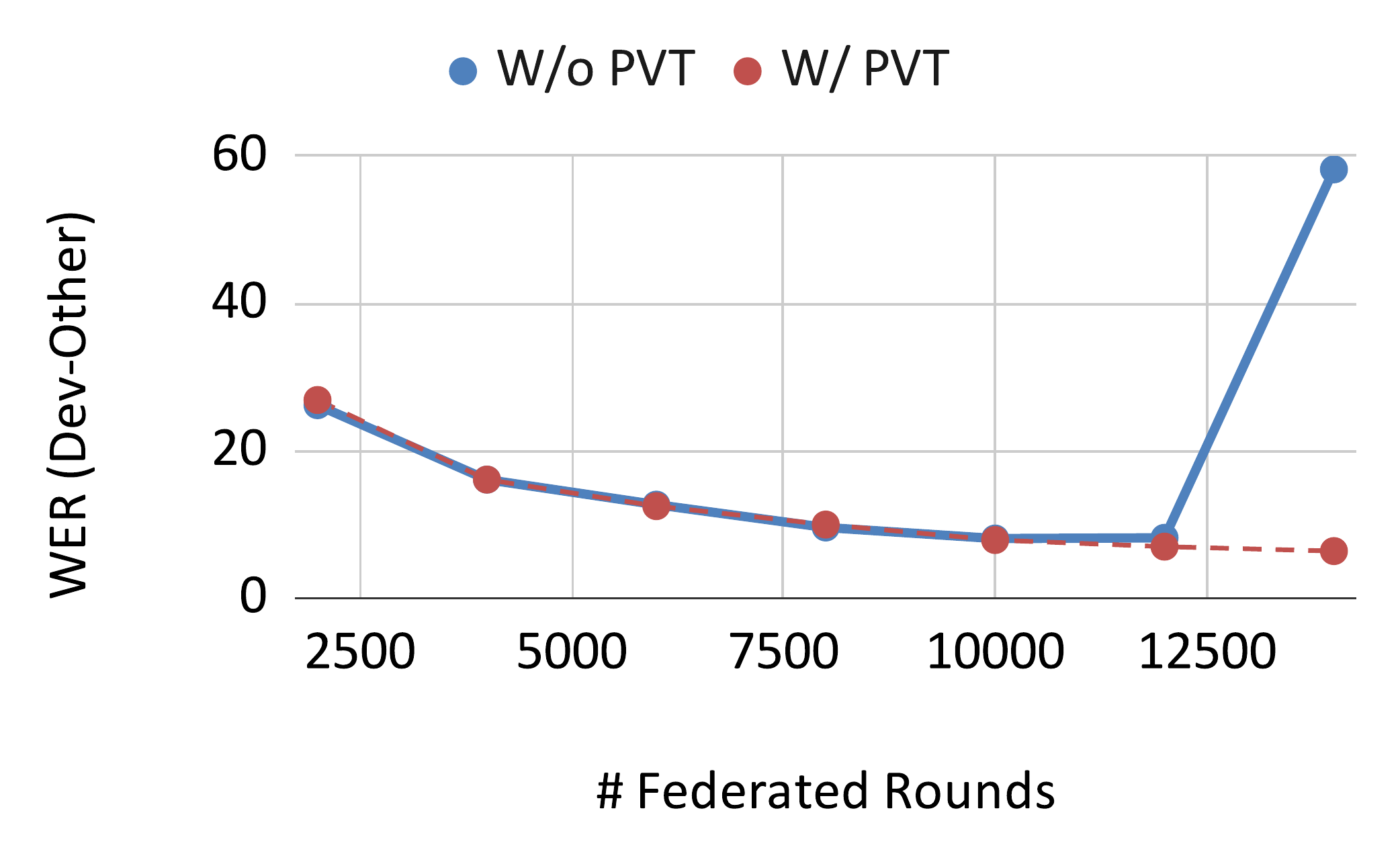}
    \vspace{-5pt}
    \caption{The comparison between with and without using per-variable transformation (PVT) when training Non-Streaming Conformer on IID LibriSpeech dataset from scratch with the S1E5M10 format.}
    \vspace{-5pt}
    \label{fig:per_variable_transformation_results}
\end{figure}

In the previous section, we showed the effectiveness of the proposed per-variable transformation for domain adaptation. We found that per-variable transformation is even more critical for from-scratch training. Fig.~\ref{fig:per_variable_transformation_results} shows its impact on WERs when we train the Non-Streaming Conformer on IID LibriSpeech dataset from scratch with the S1E5M10 format. Without applying per-variable transformation, the training is unstable. The WER first decreases and then increases after 12000 federated rounds. This issue is resolved by adding per-variable transformation, which helps stabilize training to make the WER keep decreasing.

\subsubsection{With and Without Partial Parameter Quantization}

\begin{figure}[t]
    \centering
    \includegraphics[width=0.35\textwidth]{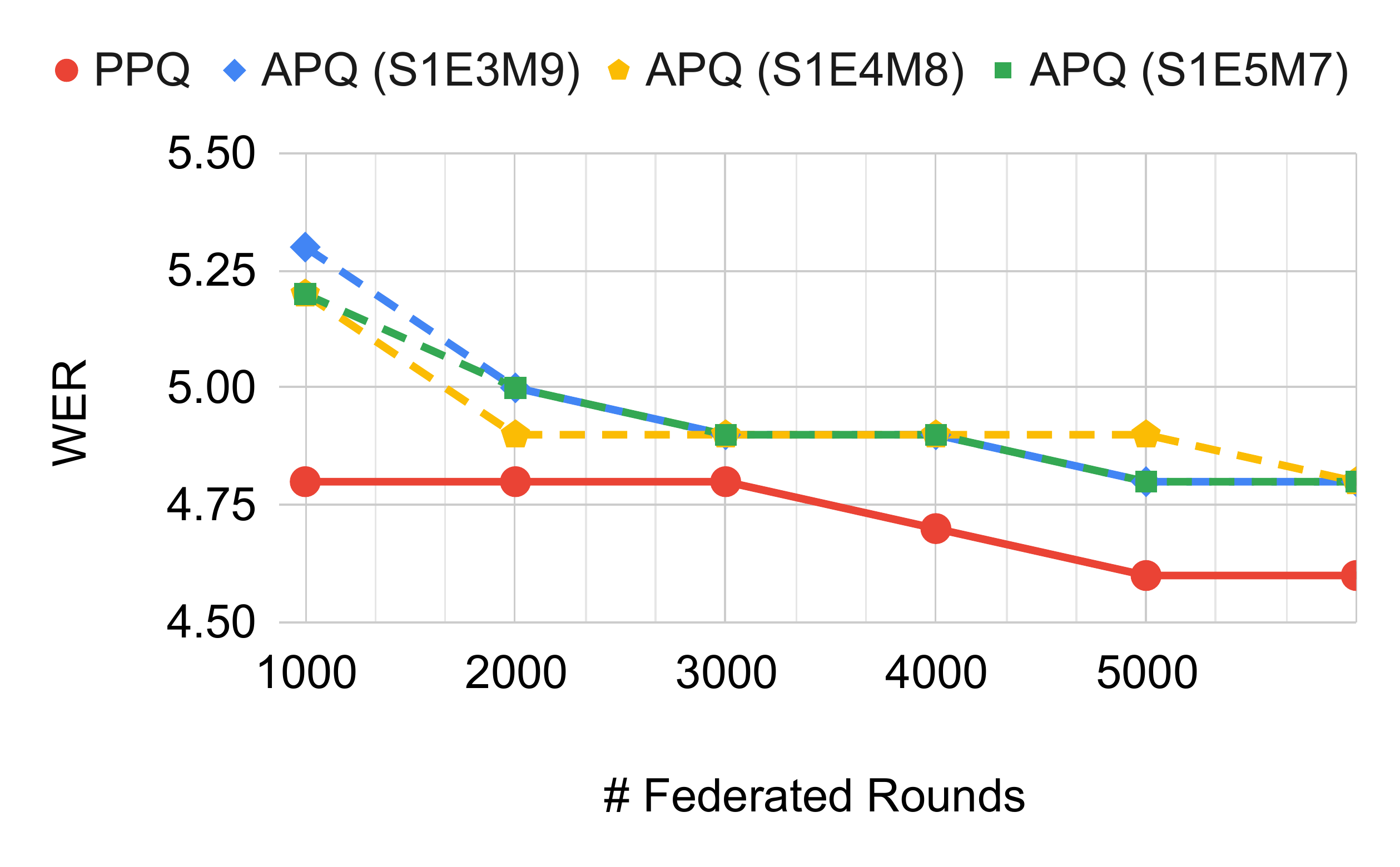}
    \vspace{-5pt}
    \caption{The comparison between partial parameter quantization (PPQ) with the 11-bit format (S1E3M7) and 90\% parameters quantized and all parameter quantization (APQ) with various 13-bit formats (S1E3M9, S1E4M8, and S1E5M7) and 100\% parameters quantized.}
    \vspace{-5pt}
    \label{fig:partial_quantization_results}
\end{figure}

In the case of quantizing 90\% parameters with the 11-bit format (S1E3M7), keeping the remaining 10\% parameters unquantized increases the average bitwidth by around 2 bits. In this study, we compare this 11-bit format with 90\% parameters quantized with various 13-bit formats with all parameters quantized. We create these 13-bit formats by allocating the extra 2 bits to the exponent and mantissa parts in different ways. These 13-bit formats are S1E3M9, S1E4M8, and S1E5M7. The training results of these formats are summarized in Fig.~\ref{fig:partial_quantization_results}. We observe that using the proposed partial parameter quantization with 11 bits results in faster convergence than all parameter quantization with 13 bits. Moreover, none of these 13-bit formats can achieve a WER as low as that of partial parameter quantization with 11 bits.

%% file: 4_related_works.tex
\section{Related Works}
\label{sec:related_works}

Most of the related works in the literature that improve inference or training efficiency focus on centralized training. One widely adopted approach is reducing the complexity of models, such as manual design~\cite{howard_2015_mobilenetv1,sandler_2018_mobilenetv2}, pruning~\cite{han_2016_deep,yang_2017_energy_pruning}, or neural architecture search~\cite{zoph_2017_nasreinforcement,yang_2021_netadaptv2}. However, reducing complexity typically limits the potential of the model for continuous improvement over growing data. Model transport compression~\cite{chraibi_2019_model_broadcast_compression} and gradient transport compression~\cite{konecny_2016_gradient_compression} keep the model unchanged and compress the transported data to save the communication cost but with the same memory usage.

Similar to OMC, Quantization-Aware Training (QAT)~\cite{rastegari_2016_xnornet,abdolrashidi_2021_quantized_resnet} also quantizes parameters. The main difference is that OMC aims to reduce memory usage \emph{during training} while QAT focuses on saving memory \emph{during inference}. When training a model, QAT stores parameters in \emph{FP32} and quantizes them on demand while OMC stores parameters in a \emph{compressed format} and decompresses them on demand. Storing parameters in FP32 allows QAT to precisely accumulate small gradients to achieve lower bitwidths for inference at the cost of no reduction in the memory usage of parameters during training. In contrast, storing parameters in a compressed format enables OMC to reduce the memory usage of parameters during training but makes it more challenging to control the quantization error and reduce bitwidths. In this paper, we propose multiple methods to effectively address this challenge.

There are a few works aiming to improve the efficiency of federated learning. Federated dropout~\cite{caldas_2018_feddrop,dhruv_2022_feddrop} trains only part of the server model on clients, so that the server model can be much more complicated than client models. However, because the client models differ from the server model, federated dropout needs to maintain a mapping between them. Similar to federate dropout, group knowledge transfer~\cite{he_2020_knowledge_transfer} also uses different models on a server and clients. The clients run a small feature extractors to extract features, which are then used to train the server model. This approach decreases client loading at the cost of increased server loading. Partial variable training~\cite{yang_2022_partial_variable_training} freezes parameters to reduce the memory usage of activations and gradients and the client-to-server communication, but the memory usage of \emph{parameters} and the \emph{server-to-client} communication are not changed. Compared to the above methods, OMC can reduce both memory usage and communication cost of parameters for federated learning without their downsides and can be further combined with them to achieve even better efficiency.

%% file: 5_conclusion.tex
\section{Conclusion}
\label{sec:conclusion}

In this paper, we proposed Online Model Compression to reduce memory usage and communication cost of model parameters for federated learning. Our realization of OMC consists of floating-point quantization, per-variable transformation, weight matrices only quantization, and partial parameter quantization. The experiments show that OMC is lightweight but can effectively maintain accuracy with significant efficiency improvement. We believe this technique will help bring state-of-the-art ASR models onto edge devices to improve user experience.

%% file: __references.bbl

%% file: 6_acknowledgements.tex
\section{Acknowledgements}

We thank Petr Zadrazil for inspiring this project and Dhruv Guliani for reviewing this paper. We also thank Sean Augenstein, Zachary Garrett and Hubert Eichner for their helpful discussions on the experiments.